\PassOptionsToPackage{numbers,sort&compress}{natbib}

\documentclass[sigconf,nonacm]{acmart}
\usepackage{graphicx}
\usepackage{soul}
\usepackage{multirow}
\usepackage{amsthm}
\usepackage{algorithm}
\usepackage{amsmath}
\usepackage{enumitem}

\AtBeginDocument{%
  }

\author{Shahroz Tariq, David Nguyen, M.A.P. Chamikara, Tingmin Wu, Alsharif Abuadbba, Kristen Moore}
\affiliation{%
  \institution{CSIRO's Data61}
  \country{Australia}
}

\begin{document}

\title{LLMs Are Not Yet Ready for Deepfake Image Detection}




\begin{abstract}
The growing sophistication of deepfakes presents substantial challenges to the integrity of media and the preservation of public trust. Concurrently, vision-language models (VLMs), large language models enhanced with visual reasoning capabilities, have emerged as promising tools across various domains, sparking interest in their applicability to deepfake detection. This study conducts a structured zero-shot evaluation of four prominent VLMs: ChatGPT, Claude, Gemini, and Grok, focusing on three primary deepfake types: faceswap, reenactment, and synthetic generation. Leveraging a meticulously assembled benchmark comprising authentic and manipulated images from diverse sources, we evaluate each model's classification accuracy and reasoning depth. Our analysis indicates that while VLMs can produce coherent explanations and detect surface-level anomalies, they are not yet dependable as standalone detection systems. We highlight critical failure modes, such as an overemphasis on stylistic elements and vulnerability to misleading visual patterns like vintage aesthetics. Nevertheless, VLMs exhibit strengths in interpretability and contextual analysis, suggesting their potential to augment human expertise in forensic workflows. These insights imply that although general-purpose models currently lack the reliability needed for autonomous deepfake detection, they hold promise as integral components in hybrid or human-in-the-loop detection frameworks.
\end{abstract}





\begin{teaserfigure}
\centering
  \includegraphics[width=1\textwidth]{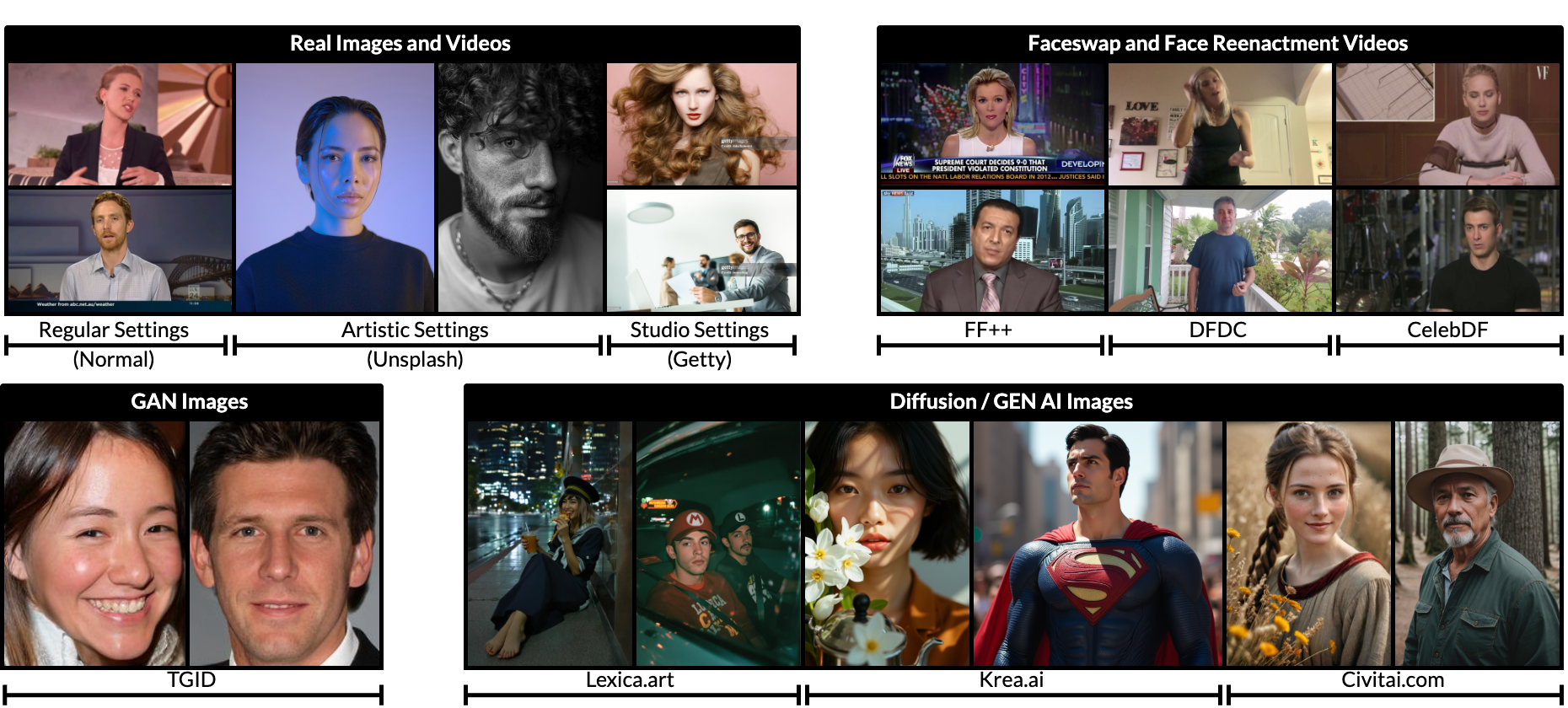}
  \caption{Evaluation Datasets. We evaluated 4 leading LLM-based vision-language models from ChatGPT, Gemini, Grok, and Claude. We handpicked samples from a diverse set of sources for both the real and fake categories and covering the three main types of deepfakes, i.e., faceswap, reenactment and synthesis.}
  \label{fig:teaser}
\end{teaserfigure}

\maketitle
\section{Introduction}
\label{sec:intro}
The rapid proliferation of deepfakes, synthetic media manipulated or generated by artificial intelligence, has introduced significant challenges to digital media authenticity~\cite{fbi2022,WDC_Metaverse,Muneer_2025, jyp}. These manipulations, often indistinguishable from real content, pose serious threats in domains such as disinformation, identity fraud, and public trust~\cite{li2022seeing,tariq2023evaluating,kim2018george,tariq2022real}. As generative models continue to evolve, so too does the sophistication of deepfakes, making their detection an increasingly urgent and complex task.

Traditionally, deepfake detection has relied on specialised computer vision models trained to identify visual artefacts or inconsistencies~\cite{woo2023qad, feng2023avad, bai2023aunet,nguyen2019capsule, CLRNet, woo2023apparatus,qian2020f3net, song2022cdnet, ble2022add, le2021exploring, TAR,lee2021detecting, ugad,wang2023altfreezing,tan2023lgrad,JeonghoPTD,MinhaCORED,MinhaFRETAL,ShallowNet1,ShallowNet2,CLRNetold,woo2023method}. While these models have achieved strong performance on benchmark datasets such as FF++ \cite{rossler2019faceforensics++}, DFDC \cite{dolhansky2020deepfake}, FakeAVCeleb \cite{HasamACMMM,khalid2021fakeavceleb} and CelebDF \cite{li2020celeb}, they often struggle to generalise to unseen manipulations or out-of-distribution data~\cite{le2025soksystematizationbenchmarkingdeepfake,cho23_RWDF,WDC_Why,Tariq2024,Tariq2024a}. Such brittleness limits their effectiveness in real-world scenarios, where deepfakes are diverse, evolving, and often adversarially crafted.

In parallel, large language models (LLMs) such as ChatGPT~\cite{OpenAI2023ChatGPT}, Gemini~\cite{Google2023Gemini}, Claude~\cite{Anthropic2023Claude}, and Grok~\cite{xAI2023Grok} have remarkable generalisation across a wide range of tasks. Recent advances in VLMs \cite{alayrac2022flamingo, li2023blip}, a class of LLMs augmented with visual reasoning capabilities, have sparked interest in their potential for media forensics~\cite{chang2023antifakeprompt}. These models can perform zero-shot inference on visual inputs using natural language prompts, raising the question: \textit{Can general-purpose VLMs serve as effective tools for deepfake detection?}

%
Despite their promise, the application of LLMs and VLMs to deepfake detection is still in its early stages.
Prior studies have typically focused on narrow subsets of deepfakes, such as synthetic images generated by GANs or diffusion models, without accounting for the broader taxonomy of manipulations, including faceswap and reenactment~\cite{jia2024can,ren2025can}. Moreover, evaluations often lack rigorous benchmarking across diverse real-world conditions, limiting the generalisability of their findings.

In this work, we present a systematic, zero-shot evaluation of four leading proprietary VLMs, ChatGPT (GPT-4o), Claude (Sonnet 4), Gemini (2.5 Flash), and Grok 3, on the task of deepfake detection. We construct a carefully curated benchmark of deepfake images spanning three core categories: faceswap, reenactment, and synthetic generation. Our dataset includes challenging real and fake samples drawn from widely-used benchmarks (FF++, DFDC, CelebDF) and generative platforms (Lexica.art, Krea.ai, Civitai.com), along with real images from regular, artistic, and studio settings. \autoref{fig:teaser} contains samples from our evaluation dataset.

Our findings reveal that while current VLMs do not match the classification accuracy of specialised vision-based detectors~\cite{CLRNet,le2025soksystematizationbenchmarkingdeepfake}, they offer valuable complementary capabilities. These include contextual reasoning, identification of perceptual cues, and the ability to generate natural-language explanations. 
%
These capabilities position VLMs not as replacements for traditional detectors but as promising collaborators in human-in-the-loop forensic workflows, enabling more interpretable and adaptable detection systems.

\noindent
\textbf{\textsc{Contributions.} } This paper makes the following key contributions:

\begin{enumerate}[leftmargin=*]
    \item \textbf{Benchmark Design:} We introduce a cross-category benchmark for evaluating VLMs on zero-shot deepfake detection, covering faceswap, reenactment, and synthesis.
    \item \textbf{Empirical Evaluation:} We provide a detailed analysis of four leading VLMs, highlighting their strengths, limitations, and failure modes across diverse real and fake image types.
    \item \textbf{Trigger Patterns:} We identified some trigger image styles for different VLMs that make them misclassify in most cases.
\end{enumerate}

\textit{We will release the dataset used in our evaluation, which includes all links, on our GitHub page once the paper is accepted.}

\section{Dataset and Experimental Setup}
\label{sec:exp}

\subsection{Datasets Construction}
To evaluate the zero-shot deepfake detection capabilities of VLMs, we curated a compact yet diverse benchmark dataset spanning three canonical deepfake categories: faceswap, reenactment, and synthetic generation, as defined by Le at al. \cite{le2025soksystematizationbenchmarkingdeepfake}. Our sampling strategy emphasised diversity in manipulation types, gender balance, and perceptual difficulty to emulate realistic detection scenarios. 

\noindent
\textbf{\textsc{Real Images}. } We collected 30 real images across three distinct settings to assess model robustness under varying visual styles: 
\begin{enumerate}[leftmargin=*]
    \item \textbf{\textit{Regular (Normal):}} Real face images (5 male, 5 female) were sampled from FF++ \cite{rossler2019faceforensics++}, DFDC \cite{dolhansky2020deepfake}, and CelebDF \cite{li2020celeb}.
    \item \textbf{\textit{Artistic (Unsplash):}} Using search terms such as ``artistic photo'' combined with gender keywords, we handpicked 10 images (5 male, 5 female) from Unsplash.
    \item \textbf{\textit{Studio (Getty):}} We retrieved 10 high-quality studio portraits (5 male, 5 female) from Getty Images using the search term ``Fashion model'' combined with gender keywords.
\end{enumerate}

This stratification allowed us to test the hypothesis that VLMs may perform well on conventional examples but struggle with more stylised or professionally curated content.


\noindent
\textbf{\textsc{Faceswap and Reenactment Deepfakes}. } We extracted video frames from three widely-used deepfake datasets, FF++ \cite{rossler2019faceforensics++}, DFDC \cite{dolhansky2020deepfake}, and CelebDF \cite{li2020celeb}, selecting 10 manipulated samples (5 male, 5 female) from each. These samples represent common manipulation techniques such as identity swapping and facial reenactment. 

\noindent
\textbf{\textsc{Synthetic Face Generation}. }
To evaluate the detection of AI-generated faces, we sourced high-quality synthetic images from:
\begin{enumerate}[leftmargin=*]
    \item \textbf{\textit{GAN-based:}} 10 samples (5 male, 5 female) from the R3GAN dataset~\cite{huang2024gan}, a recent GAN benchmark (NeurIPS '24).
    \item \textbf{\textit{Diffusion-based:}} 30 samples (15 male, 15 female) from Lexica.art, Krea.ai, and Civitai.com, using gendered search terms.
\end{enumerate}

\noindent
\textbf{\textsc{Difficulty and Edge Cases.} } To stress-test model performance, we included samples with challenging visual attributes such as occlusion, complex lighting, and hyper-realistic rendering. These edge cases were selected to probe the limits of VLM reasoning and highlight failure modes. 

\subsection{Model Selection and Prompting Protocol}
We evaluated four leading proprietary VLMs: ChatGPT (GPT-4o), Gemini (2.5 Flash), Grok 3, and Claude (Sonnet 4). All models were accessed via their respective web interfaces and prompted in a zero-shot setting, without any task-specific fine-tuning or in-context examples. 

\noindent
\textbf{\textsc{Prompting Strategy.} } 
To ensure consistency across models, we used a unified persona prompt (link anonymised for double-blind review) followed by a structured task prompt for each image: \textit{``Analyse the following image. Follow your analysis framework.''}. Models were instructed to produce responses in the following format:
\begin{itemize}[leftmargin=*]
    \item \textbf{Decision:} [Real / Fake / Inconclusive]
    \item \textbf{Confidence:} [e.g., 82\% confident it's real]
    \item \textbf{Reasoning:} [Detailed paragraph with evidence and rationale]
    \item \textbf{Uncertainty:} [If confidence < 80\%, describe limitations]
\end{itemize}
This format was designed to elicit interpretable outputs and facilitate comparative analysis across models.


\noindent
\textbf{\textsc{Evaluation Procedure}. }
Each image was independently assessed by all four VLMs. We recorded the predicted class, confidence score, and explanation. To quantify performance, we computed standard classification metrics, accuracy, precision, recall, and F1-score, for each model across different image categories. All metrics were scaled between 0 and 1. Given the probabilistic nature of LLMs, we also monitored consistency across repeated queries for a subset of samples. In some cases, we observed variability in both confidence scores and predicted labels for the same image, highlighting the non-deterministic behaviour of these models (see \autoref{sec:dis}).

\section{Results}
\label{sec:res}

\subsection{Quantitative Analysis}
We evaluate the performance of four VLMs across different categories of real and deepfakes images and report standard classification metrics for each model in \autoref{tab:real} and \autoref{tab:faceswap_reenact_synthesis}.

\noindent
\textbf{\textsc{How well the VLMs perform on Real Images?} }
\autoref{tab:real} presents results on real images across three settings: regular (Normal), artistic (Unsplash), and studio-quality (Getty). All models perform perfectly on regular images, but their accuracy degrades as the images become more artistic and polished (\textit{normal} $\rightarrow$ \textit{unsplash} $\rightarrow$ \textit{getty}).

\begin{itemize}[leftmargin=*]
    \item ChatGPT and Claude show strong performance on artistic images but drop in accuracy on studio-quality images, suggesting sensitivity to hyper-realistic aesthetics.
    \item Grok performs well only on regular images and fails to generalise to more stylised content.
    \item Gemini maintains high accuracy across all real image types, but as we show later, this is due to a strong bias toward predicting the ``real'' class, even for fake inputs.
\end{itemize}

\textbf{Insight:} These results highlight a key limitation: VLMs may overfit to surface-level realism cues, misclassifying high-quality real images as fake or vice versa. This undermines their reliability in real-world scenarios where image quality varies widely.


%
\begin{table}
\centering
\caption{The performance on real images and frames from real videos. We used three settings for the real content.}
\label{tab:real}
\resizebox{\linewidth}{!}{%
\begin{tabular}{l|ccc|ccc|ccc|r} 
\hline
\multirow{3}{*}{\textbf{Method}} & \multicolumn{10}{c}{\textbf{\textsc{Real Images and Videos}}} \\ 
\cline{2-11}
 & \multicolumn{3}{c|}{\begin{tabular}[c]{@{}c@{}}\textbf{Regular Settings}\\\textbf{(Normal)}\end{tabular}} &
 \multicolumn{3}{c|}{\begin{tabular}[c]{@{}c@{}}\textbf{Artistic Settings}\\\textbf{(Unsplash)}\end{tabular}} & 
 \multicolumn{3}{c|}{\begin{tabular}[c]{@{}c@{}}\textbf{Studio Settings}\\\textbf{(Getty)}\end{tabular}} &
 \multicolumn{1}{c}{\multirow{2}{*}{\begin{tabular}[c]{@{}c@{}}Avg.\\Acc.\end{tabular}}} \\ 
\cline{2-10}
 & \textit{Pre.} & \textit{Rec.} & \textit{F1} & \textit{Pre.} & \textit{Rec.} & \textit{F1} & \textit{Pre.} & \textit{Rec.} & \textit{F1} & \multicolumn{1}{c}{} \\ 

\hline
\textbf{Grok} & 1.00 & 1.00 & 1.00 & 1.00 & 0.50 & 0.67 & 1.00 & 0.20 & 0.33 & 0.57 \\ 
\hline
\textbf{Claude} & 1.00 & 1.00 & 1.00 & 1.00 & 0.80 & 0.89 & 1.00 & 0.50 & 0.67 & 0.77 \\
\hline
\textbf{ChatGPT} & 1.00 & 1.00 & 1.00 & 1.00 & 0.80 & 0.89 & 1.00 & 0.70 & 0.82  & 0.83 \\ 
\hline
\textbf{Gemini} & 1.00 & 0.70 & 0.82 & 1.00 & 1.00 & 1.00 & 1.00 & 1.00 & 1.00 & 0.90 \\ 
\hline
\end{tabular}
}
\end{table}
\noindent
\textbf{\textsc{How well do VLMs perform on Deepfakes?} }
\autoref{tab:faceswap_reenact_synthesis} summarises model performance on three deepfake categories: faceswap, reenactment, and synthetic generation (GANs and diffusion-based).

\begin{itemize}[leftmargin=*]
    \item ChatGPT consistently outperforms other models across all deepfake types, achieving the highest average accuracy and F1-scores.
    \item Claude shows moderate performance, particularly on GANs and diffusion images, but struggles with faceswap and reenactment.
    \item Gemini and Grok perform poorly across the board, with Grok failing entirely on faceswap and GANs (0\% accuracy).
    \item Notably, Gemini achieves high precision on real images but misclassifies most fake images as real, indicating a skewed decision boundary.
\end{itemize}

\textbf{Insight:} While ChatGPT demonstrates relatively better generalisation, none of the models achieve robust performance across all deepfake types. This reinforces the need for task-specific training or hybrid detection pipelines.



\begin{table*}
\centering
\caption{The performance on typical deepfakes, i.e., faceswap and face reenactment, and on synthesis deepfakes, i.e., diffusion / generative AI  and GANs. We used samples from three famous deepfake benchmarking datasets, one source for GAN, and three sources for diffusion / generative AI images. }
\label{tab:faceswap_reenact_synthesis}
\resizebox{\linewidth}{!}{%
\begin{tabular}{l|ccc|ccc|ccc|c|cccc|ccc|ccc|ccc|c} 
\hline
\multirow{4}{*}{\textbf{Method}} & \multicolumn{10}{c|}{\textbf{\textsc{Faceswap and Face Reenactment}}} & \multicolumn{14}{c}{\textbf{\textsc{Synthesis}}} \\ 
\cline{2-25}
 & \multicolumn{3}{c|}{\multirow{2}{*}{\textbf{\textit{FF++}}}} & \multicolumn{3}{c|}{\multirow{2}{*}{\textbf{DFDC}}} & \multicolumn{3}{c|}{\multirow{2}{*}{\textbf{CelebDF}}} & \multirow{3}{*}{\begin{tabular}[c]{@{}c@{}}\textit{Avg.}\\\textit{Acc.}\end{tabular}} & \multicolumn{4}{c|}{\textbf{GAN}} & \multicolumn{10}{c}{\textbf{Diffusion / GenAI}} \\ 
\cline{12-25}
 & \multicolumn{3}{c|}{} & \multicolumn{3}{c|}{} & \multicolumn{3}{c|}{} &  & \multicolumn{4}{c|}{\textbf{\textit{TGID}}} & \multicolumn{3}{c|}{\textbf{\textit{lexica.art}}} & \multicolumn{3}{c|}{\textbf{\textit{Krea.ai}}} & \multicolumn{3}{c|}{\textbf{Civitai.com}} & \multirow{2}{*}{\begin{tabular}[c]{@{}c@{}}\textit{Avg.}\\\textit{Acc.}\end{tabular}} \\ 
\cline{2-10}\cline{12-24}
 & \textit{Pre.} & \textit{Rec.} & \textit{F1} & \textit{Pre.} & \textit{Rec.} & \textit{F1} & \textit{Pre.} & \textit{Rec.} & \textit{F1} &  & \textit{Pre.} & \textit{Rec.} & \textit{F1} & \textit{Acc.} & \textit{Pre.} & \textit{Rec.} & \textit{F1} & \textit{Pre.} & \textit{Rec.} & \textit{F1} & \textit{Pre.} & \textit{Rec.} & \textit{F1} &  \\ 
\hline
\textbf{Grok} & 0.00 & 0.00 & 0.00 & 0.00 & 0.00 & 0.00 & 0.00 & 0.00 & 0.00 & 0.00 & 0.00 & 0.00 & 0.00 & 0.00 & 1.00 & 0.40 & 0.57 & 1.00 & 0.30 & 0.46 & 1.00 & 0.10 & 0.18 & 0.27 \\ 
\hline
\textbf{Gemini} & 1.00 & 0.30 & 0.46 & 0.00 & 0.00 & 0.00 & 0.00 & 0.00 & 0.00 & 0.10 & 1.00 & 0.50 & 0.67 & 0.50 & 0.00 & 0.00 & 0.00 & 1.00 & 0.40 & 0.57 & 1.00 & 0.40 & 0.57 & 0.27 \\ 
\hline
\textbf{Claude} & 1.00 & 0.50 & 0.67 & 1.00 & 0.10 & 0.18 & 1.00 & 0.30 & 0.46 & 0.30 & 1.00 & 0.30 & 0.46 & 0.30 & 1.00 & 0.30 & 0.46 & 1.00 & 0.80 & 0.89 & 1.00 & 0.70 & 0.82 & 0.60 \\
\hline
\textbf{ChatGPT} & 1.00 & 0.60 & 0.75 & 1.00 & 0.80 & 0.89 & 1.00 & 0.90 & 0.95 & 0.77 & 1.00 & 1.00 & 1.00 & 1.00 & 1.00 & 0.50 & 0.67 & 1.00 & 0.80 & 0.89 & 1.00 & 0.70 & 0.82 & 0.67 \\
\hline

\end{tabular}
}
\end{table*}


\subsection{Qualitative Analysis}
Beyond raw performance metrics, we conducted a qualitative analysis of model outputs to better understand their reasoning patterns, failure modes, and potential for human-AI collaboration.

\begin{figure}
    \centering
    \includegraphics[width=1\linewidth]{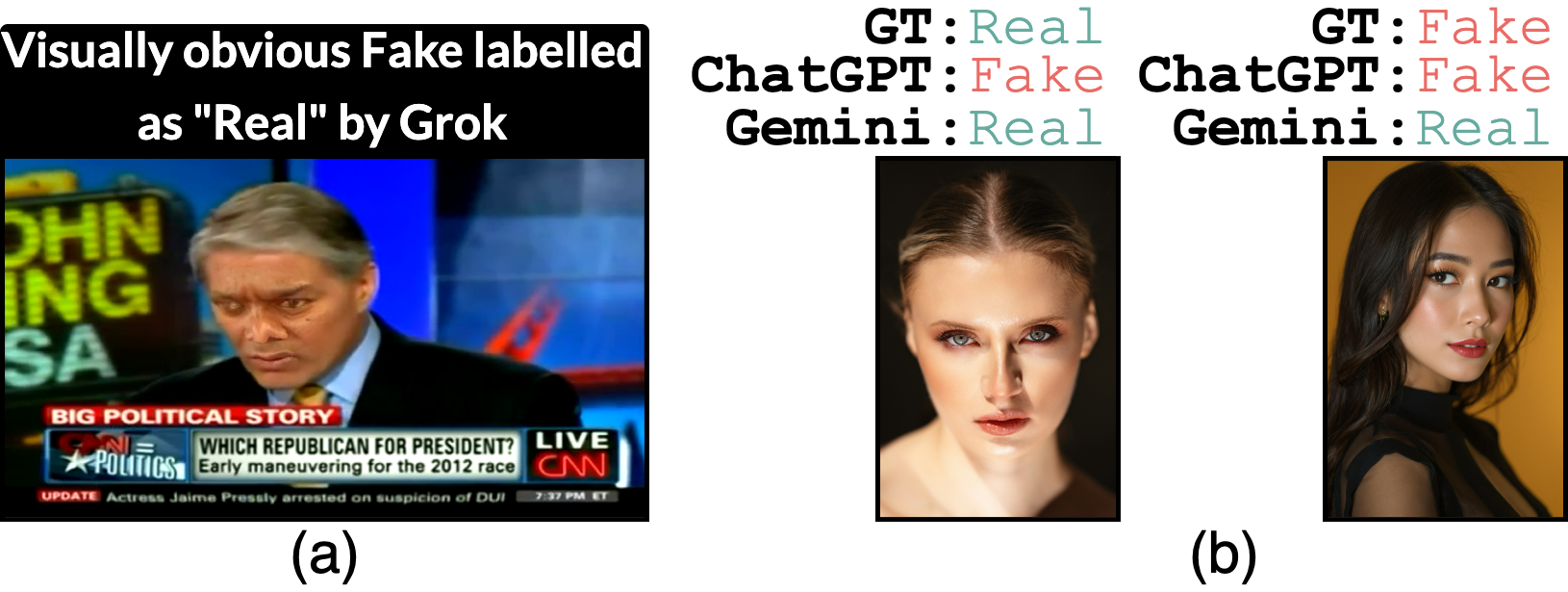}
    \caption{Failure cases of different models. Here, GT denotes ground truth.}
    \label{fig:fails}
    \vspace{-8pt}
\end{figure}

\noindent
\textbf{\textsc{Failure Modes and Misclassification Patterns}. }
Our analysis revealed several recurring failure patterns across models:
\begin{itemize}[leftmargin=*]
    \item \textbf{\textit{Surface-level reasoning:}} Grok frequently misclassified manipulated images as real due to over-reliance on background elements or text overlays, which appeared realistic. In one extreme case (\autoref{fig:fails}a), a visibly distorted face was labelled as real because the surrounding context was convincing.
    \item \textbf{\textit{Hyper-realism confusion:}} Gemini consistently labelled hyper-realistic diffusion images as real, failing to detect subtle artefacts. This same bias led it to correctly classify studio-quality real images, but also to misclassify similarly styled fakes (\autoref{fig:fails}b).
    \item \textbf{\textit{Overcautious behaviour:}} ChatGPT, while generally more accurate, occasionally produced false positives, labelling real images as fake, especially in artistic or studio settings. This suggests a conservative bias when encountering unfamiliar or stylised content (\autoref{fig:fails}b).
\end{itemize}

These observations underscore the models' reliance on superficial visual cues rather than deep semantic understanding of facial integrity or manipulation artefacts.

\begin{figure}
    \centering
    \includegraphics[width=1\linewidth]{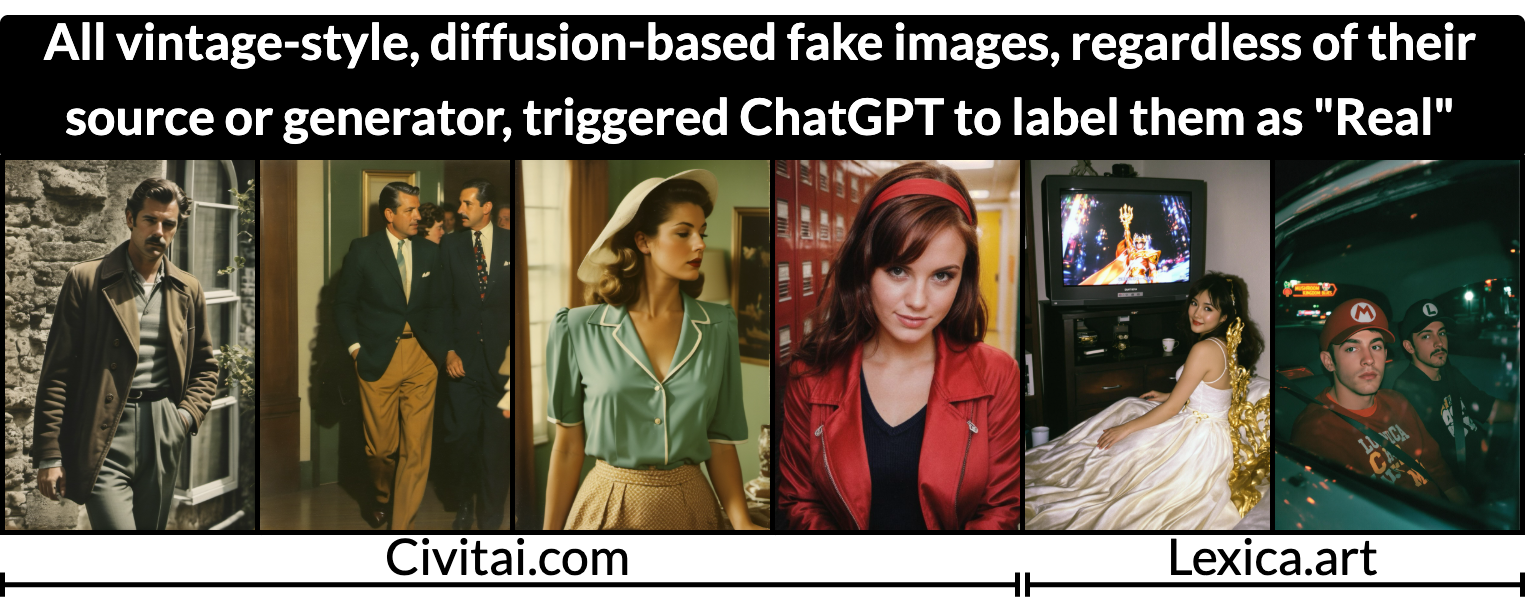}
    \caption{Trigger images for ChatGPT.}
    \label{fig:Trigger}
    \vspace{-15pt}
\end{figure}

\noindent
\textbf{\textsc{Trigger Patterns and Style Biases.} } A particularly interesting finding emerged with ChatGPT (GPT-4o): vintage-style diffusion images were consistently misclassified as real. To probe this further, we collected additional vintage-style samples from Lexica and Civitai (\autoref{fig:Trigger}), all of which were also labelled as real. This suggests a potential style bias, where certain aesthetic features (e.g., sepia tones, film grain) act as implicit priors for authenticity in the model's internal representation.

\noindent
\textbf{\textsc{Interpretability and Explanation Quality.} }
Despite their classification limitations, VLMs demonstrated notable strengths in generating natural-language rationales. These explanations often included references to facial symmetry, lighting inconsistencies, or texture anomalies. While not always accurate, they provide a valuable interpretive layer that could assist human analysts in ambiguous cases. Claude and ChatGPT produced the most coherent and detailed justifications. Grok explanations were often vague or misdirected (e.g., focusing on irrelevant image regions). Gemini explanations tended to be overly confident, even when incorrect.

\textbf{Insight:} These findings support the notion that VLMs, while unreliable as standalone detectors, can serve as interpretability tools, surfacing cues, raising uncertainty, and supporting human decision-making in forensic workflows.

\section{Discussion}
\label{sec:dis}

\subsection{Key Takeaways and Implications}
Our evaluation shows that while current VLMs can generate plausible explanations and pick up on surface-level cues, their outputs are not yet reliable enough to serve as a strong signal for deepfake detection.
Their performance is inconsistent across manipulation types and real image styles, and they are prone to both false positives and false negatives depending on visual context.

\noindent
\textbf{\textsc{Human-AI Collaboration: A More Viable Path.} }
Despite limitations in classification accuracy, VLMs exhibit strengths that position them as promising collaborative agents in forensic workflows~\cite{tariq2025a2c,irons2024towards}. Their interpretability, complementary capabilities, and accessibility suggest that VLMs are better suited as assistive tools rather than autonomous detectors, helping bridge the gap between technical outputs and human reasoning\cite{tariq2025bridging}.


\noindent
\textbf{\textsc{Model Behaviour and Ethical Considerations.} } The observed biases, such as ChatGPT's tendency to misclassify vintage-style images or Gemini's inclination to overpredict the ``real'' class, raise important concerns about model transparency and trustworthiness. In high-stakes applications like misinformation detection or legal evidence review, such biases could lead to serious consequences. Moreover, the non-deterministic nature of LLMs introduces variability in outputs, even for identical inputs. This stochasticity complicates reproducibility and accountability, especially when models are deployed in real-world decision-making pipelines.

\subsection{Limitations and Future Directions}

\noindent
\textbf{\textsc{Limitations.} } While our study provides valuable insights into the capabilities and shortcomings of current VLMs for deepfake detection, several limitations must be acknowledged:

\begin{itemize}[leftmargin=*]
    \item \textbf{\textit{Dataset Scale:}} Our benchmark comprises 100 curated samples, allowing for detailed manual analysis but limiting statistical generalisability. Larger-scale evaluations are needed to validate these findings across broader distributions.
    \item \textbf{\textit{Model Scope:}} We focused on four proprietary VLMs. Open-source alternatives (e.g., LLaVA~\cite{liu2023visual}, MiniGPT-4~\cite{zhuminigpt}) were not included but may offer different trade-offs in performance, transparency, and customisability.
    \item \textbf{\textit{Image-Only Evaluation:}} Our analysis was restricted to static images. In practice, deepfakes often appear in video form, where temporal inconsistencies and motion artefacts provide additional detection cues.
    \item \textbf{\textit{Prompt Sensitivity:}} The performance of VLMs is highly sensitive to prompt phrasing. While we used a consistent and structured prompt, alternative formulations may yield different results.
\end{itemize}

\noindent
\textbf{\textsc{Future Directions.} } Building on our findings, we outline several promising avenues for future research:

\begin{itemize}[leftmargin=*]
    \item \textbf{\textit{Multimodal Benchmarking:}} Expanding evaluations to include video sequences, audio-visual manipulations, and adversarially crafted fakes will provide a more comprehensive stress test of VLM capabilities.
    \item \textbf{\textit{Hybrid Architectures:}} Integrating VLMs with specialised vision encoders or forensic feature extractors may improve grounding and reduce hallucinations, enabling more accurate and explainable detection.
    \item \textbf{\textit{Open-Source Model Evaluation:}} Assessing the performance of open-source VLMs can democratise research in this space and facilitate reproducibility, fine-tuning, and community-driven improvements.
    \item \textbf{\textit{Interactive Forensics Tools:}} Embedding VLMs within user-facing interfaces, such as dashboards for journalists or moderation tools for platforms, can enhance transparency and support real-time decision-making.
\end{itemize}

\section{Conclusion}
This paper presented a focused evaluation of state-of-the-art VLMs, ChatGPT, Claude, Gemini, and Grok, on the task of zero-shot deepfake detection. Using a carefully curated benchmark spanning faceswap, reenactment, and synthetic generation, we demonstrated that while these models exhibit impressive generalisation and interpretability, they fall short as reliable standalone deepfake detectors.

Our findings reveal that VLMs are prone to misclassification, particularly when faced with hyper-realistic or stylistically ambiguous content. However, their ability to generate contextual explanations and surface perceptual cues suggests a promising role as collaborative agents in human-AI forensic pipelines.

We argue that the future of deepfake detection lies not in replacing specialised detectors with general-purpose models, but in designing hybrid systems that combine the strengths of both. By embedding VLMs within interactive, explainable, and human-centred workflows, we can enhance trust, transparency, and robustness in the fight against synthetic media.

\bibliographystyle{unsrt}
\balance
\bibliography{references}

\end{document}